\documentclass[11pt]{article}

\usepackage[final]{acl}

\usepackage{times}
\usepackage{latexsym}
\usepackage{comment} 
\usepackage{todonotes}
\usepackage{enumitem}
\usepackage{makecell}
\usepackage[most]{tcolorbox}
\usepackage{float}
\usepackage{listings}
\usepackage{booktabs}
\usepackage{arydshln}
\lstdefinestyle{promptstyle}{
   basicstyle=\ttfamily\fontsize{9pt}{10pt}\selectfont,
   breaklines=true,
   breakatwhitespace=false,
   breakautoindent=true,
   columns=fullflexible,
   keepspaces=true,
   showstringspaces=false,
   xleftmargin=0pt,
   xrightmargin=0pt,
   literate={\{}{{\{}}1
         {\}}{{\}}}1
         {[}{{[}}1
         {]}{{]}}1
         {:}{{:}}1
         {,}{{,}}1
}
\usepackage{arydshln}
\usepackage{placeins}
\usepackage[T1,T5]{fontenc}

\usepackage[utf8]{inputenc}

\usepackage{microtype}

\usepackage{inconsolata}

\usepackage{graphicx}
\usepackage[table]{xcolor}
\usepackage{multirow}
\usepackage{tabularx}
\usepackage{array}
\DeclareUnicodeCharacter{1ED9}{\d{\^o}}      
\DeclareUnicodeCharacter{1EAD}{\d{\^a}}      
\DeclareUnicodeCharacter{1EC7}{\d{\^e}}      
\DeclareUnicodeCharacter{01B0}{\uhorn}        
\DeclareUnicodeCharacter{0111}{\dj}          
\DeclareUnicodeCharacter{00F4}{\^o}          
\DeclareUnicodeCharacter{00F0}{\dh}          
\DeclareUnicodeCharacter{01A1}{\ohorn}       
\DeclareUnicodeCharacter{1EC3}{\d{\~e}}      
\DeclareUnicodeCharacter{1EBF}{\'{\^e}}      
\DeclareUnicodeCharacter{1EC1}{\`{\^e}}      
\DeclareUnicodeCharacter{1EE7}{\h{u}}        
\DeclareUnicodeCharacter{1EC5}{\~{\^e}}      

\title{En-ViMedNER: An English-Vietnamese Parallel Biomedical Corpus\\with UMLS Semantic Type Annotations}

\author{
  \textbf{Nhu Vo\textsuperscript{1,2}}\thanks{Equal contribution.},
  \textbf{Phuong Nguyen\textsuperscript{1}}\footnotemark[1],
  \textbf{Nu Uyen Phuong Le\textsuperscript{1}},
  \textbf{I\~nigo Jauregi Unanue\textsuperscript{2}},
\\
  \textbf{Dung D. Le\textsuperscript{1,3}},
  \textbf{Massimo Piccardi\textsuperscript{2}},
  \textbf{Wray Buntine\textsuperscript{1,4}}
\\
  \textsuperscript{1}College of Engineering and Computer Science, VinUniversity, Vietnam,
\\
  \textsuperscript{2}Faculty of Engineering and IT, University of Technology Sydney, Australia,
\\
  \textsuperscript{3}Center for AI Research, VinUniversity, Vietnam,
\\
  \textsuperscript{4}Monash University, Australia
\\
  \small{
   \textbf{Correspondence:} \href{mailto:nhu.vd@vinuni.edu.vn}{nhu.vd@vinuni.edu.vn}
  }
}

\begin{document}
\maketitle
\begin{abstract}
Biomedical Named Entity Recognition (NER) is fundamental to healthcare AI applications, including clinical decision support and medical information extraction. While corpora with Unified Medical Language System (UMLS) annotations, such as MedMentions, have driven progress in English biomedical NER, no comparable resource exists for Vietnamese. This paper presents En-ViMedNER, the first English-Vietnamese parallel biomedical NER corpus annotated with UMLS semantic types, which are language-neutral codes providing a shared cross-lingual label space and ensuring direct comparability with existing UMLS-based resources. The corpus contains 4,392 PubMed abstract pairs, 44,892 English-Vietnamese sentence pairs, and 202,949 aligned entity-mention pairs across 21 semantic types adapted from the MedMentions ST21pv dataset. To balance quality and scalability, we have constructed the corpus through automatic translation, expert post-editing, LLM-assisted label projection, and human verification and adjudication. We characterize En-ViMedNER as a large-scale silver-standard corpus with a human-audited and consensus-corrected mini-test subset. We evaluate En-ViMedNER in two settings: (i) Vietnamese-input/Vietnamese-output biomedical NER and (ii) English-input/Vietnamese-output cross-lingual NER. For Vietnamese NER, we benchmark Vietnamese-supervised encoder models, English-supervised multilingual encoder models, and prompt-based LLMs. The best model achieves an F1 score of 52.70 on the test set and 53.78 on the mini-test set. For cross-lingual NER, we benchmark encoder-decoder models and prompt-based LLMs. The best model achieves an F1 score of 45.44 on the mini-test set. We publicly release our corpus, corpus construction pipeline, and baseline models to facilitate future Vietnamese biomedical NLP research\footnote{Our En-ViMedNER dataset and baseline fine-tuned models are publicly available at: \texttt{\url{https://huggingface.co/collections/nhuvo/en-vimedner}}.}. Given the semi-automated approach and the positive use cases of the corpus, we argue similar efforts should be made for some other low-resource languages.

\end{abstract}

\section{Introduction}

Biomedical named entity recognition (NER) is a core task in biomedical information extraction (IE) \citep{perera2020biomedical}. It identifies mentions of biomedical concepts in unstructured text and supports downstream applications such as biomedical knowledge extraction, information retrieval, and question answering. While English biomedical NLP has benefited from large annotated resources and language models trained on biomedical text, Vietnamese biomedical NLP remains under-resourced, especially for entity recognition grounded in biomedical ontologies \citep{lee2020biobert, phan2022vietbioner}.

Previous work has shown that biomedical NER depends on both annotated corpora and biomedical knowledge resources. UMLS is a widely used biomedical ontology that provides semantic types for biomedical concepts \citep{bodenreider2004unified}. MedMentions extended this line of work by linking PubMed abstracts to UMLS concepts and semantic types at scale \citep{mohan2019medmentions}. In parallel, BioBERT showed that pretraining on biomedical text can improve biomedical NER and related text mining tasks \citep{lee2020biobert}.

In Vietnamese, recent work has started to expand resources for biomedical and healthcare NLP, including PhoNER\_COVID19 \citep{truong2021phoner}, ViHealthBERT \citep{nguyen2022vihealthbert}, VietBioNER \citep{phan2022vietbioner}, ViMedNER \citep{duong2024vimedner}, and VietMedNER \citep{leduc2025vietmedner}. However, these resources are still limited in scale, number of entity types, and are designed for spoken data or task-specific label sets, rather than a broad ontology such as UMLS.

This leaves two related gaps. First, to the best of our knowledge, there is still no publicly described English-Vietnamese biomedical NER resource that preserves UMLS semantic annotations, which are provided in English, for Vietnamese text. As a result, current Vietnamese biomedical NER research cannot benefit from cross-lingual transfer from English biomedical resources and interoperability with biomedical knowledge systems. Second, because biomedical documents and established biomedical NER datasets/corpora are much more abundant in English than in Vietnamese, it is important to examine whether English biomedical supervision can support Vietnamese biomedical NER and English-to-Vietnamese cross-lingual NER, where a model converts the input English sentence into a Vietnamese sentence with NER tags. This setting is challenging because models must not only translate biomedical text, but also preserve entity boundaries and semantic labels across languages.

In this paper, we present \textbf{En-ViMedNER}, the first English-Vietnamese parallel biomedical NER corpus with UMLS-based semantic type annotations for Vietnamese text. Unlike existing Vietnamese biomedical NER resources, En-ViMedNER projects and preserves the semantic type annotations from MM-ST21pv\footnote{MM-ST21pv contains annotations of 21 semantic types \citep{mohan2019medmentions}, while the full MedMentions corpus contains annotations of 126 semantic types \citep{fraser2019extracting}.} across aligned English-Vietnamese sentence pairs. Built from MedMentions through automatic translation, expert post-editing, LLM-assisted label projection, and manual verification and adjudication, the final corpus contains 4,392 PubMed abstract pairs, 44,892 English-Vietnamese sentence pairs, and 202,949 aligned entity-mention pairs. En-ViMedNER should be viewed as a large-scale silver-standard resource with exhaustive independent human verification and consensus adjudication is limited to the 450-sentence mini-test subset. En-ViMedNER supports both Vietnamese biomedical NER and English-input/Vietnamese-output cross-lingual NER.

Our main contributions are as follows:
\begin{itemize}
   \item We release En-ViMedNER, a parallel English-Vietnamese biomedical NER corpus with UMLS-based Vietnamese annotations.

   \item We provide comprehensive benchmarks for Vietnamese biomedical NER, comparing Vietnamese-supervised encoder fine-tuning, English-supervised multilingual encoder fine-tuning, and prompt-based LLMs.

   \item We evaluate English-input/Vietnamese-output cross-lingual biomedical NER using fine-tuned encoder-decoder models and prompt-based LLMs, and compare direct end-to-end generation with cascaded translation-and-NER pipelines.

   \item We introduce an LLM-as-a-judge evaluation protocol for cross-lingual NER generation, enabling more robust assessment of entity correctness under cross-lingual semantic variation.

   \item We publicly release the corpus, corpus construction pipeline, and benchmark setup to support reproducible research in Vietnamese biomedical NLP.
\end{itemize}
Although this work focuses on Vietnamese, the pipeline can be adapted to construct UMLS-based biomedical NER corpora for other low-resource languages, provided that reliable biomedical MT, target-language medical review, and span validation are available.
\section{Background and Related Work}
In this section, we introduce the key concepts and resources underlying our work. 

\subsection{UMLS and Semantic Types}
\label{subsec:umls-and-the-semantic-network}
The Unified Medical Language System (UMLS), maintained by the U.S. National Library of Medicine \citep{D.A.B.Lindberg:93}, is a comprehensive biomedical knowledge resource that integrates over 200 source vocabularies\footnote{\url{https://uts.nlm.nih.gov/uts/umls/home}} to facilitate interoperability across biomedical information systems. One of the three UMLS knowledge sources is the Semantic Network. The Semantic Network is composed of 127 semantic types as nodes and 54 semantic relations as links between those nodes\footnote{\url{https://www.ncbi.nlm.nih.gov/books/NBK9679/}}. Semantic types enable standardized classification of all biomedical concepts in the UMLS Metathesaurus, making them well-suited as label categories for biomedical NER.

\subsection{MedMentions Corpus}
\label{subsec:medmentions-corpus}
MedMentions \citep{mohan2019medmentions} is a large-scale biomedical corpus designed to promote research in biomedical named entity recognition and concept normalization. The corpus consists of 4,392 PubMed abstracts randomly selected from papers published in 2016. Each abstract was manually annotated by professional experts who identified entity mentions and linked them to UMLS concepts using Concept Unique Identifiers (CUIs). In total, the corpus contains 352,496 mentions corresponding to 34,724 unique UMLS concepts spanning 126 of 127 semantic types, with a reported inter-annotator agreement of 97.3\%.

In addition to the full corpus, \citeauthor{mohan2019medmentions} released MedMentions ST21pv (MM-ST21pv), a filtered version retaining only mentions mapped to the 21 clinically relevant semantic types in the ST21pv subset across 18 source ontologies. This subset provides a more tractable label space while maintaining broad biomedical coverage. Our work derives from MM-ST21pv, translating the English abstracts into Vietnamese and projecting the semantic type annotations onto the translated text.

\subsection{Vietnamese Biomedical NER}
\label{subsec:vietnamese-biomedial-ner}
Research on Vietnamese biomedical NER remains limited compared to English. \citet{Ngo2019} constructed a Vietnamese medical terminology resource integrable into UMLS, mapping approximately 11,383 disease terms and 7,146 drug names to English corresponding to UMLS concepts. While valuable for lexical lookup, this resource does not provide annotated text for training NER models.

More recently, \citet{duong2024vimedner} introduced a Vietnamese medical NER dataset named ViMedNER with initial benchmarks using pre-trained language models. \citet{leduc2025vietmedner} presented a greater effort with 18 entity types across 9,000 sentences of medical speech transcripts in a dataset named VietMedNER. Their experiments demonstrated that encoder-only architectures (BERT, XLM-R, DeBERTa) consistently outperform sequence-to-sequence models (BART, T5) for Vietnamese medical NER. Notably, VietMedNER employed a large test set relative to the training set to obtain more statistically significant evaluation results when leveraging large pre-trained models.

However, no existing Vietnamese biomedical NER corpus provides annotations with UMLS semantic types. Our work addresses this gap by constructing the first such resource derived from MedMentions ST21pv.

\subsection{Cross-Lingual Transfer} 
\label{subsection:cross-lingual} 

Cross-lingual transfer is a practical strategy for biomedical NER because high-quality annotations are often concentrated in English, while target languages such as Vietnamese remain under-resourced \citep{miftahutdinov2020biomedical}. Prior studies show that multilingual encoders such as mBERT and XLM-R can support cross-lingual transfer \citep{pires2019multilingual,conneau2020unsupervised}, including in biomedical and clinical NER \citep{hakala-pyysalo-2019-biomedical,miftahutdinov2020biomedical}. However, transfer performance can be affected by language distance, domain mismatch, terminology variation, and entity-boundary differences \citep{rivera2021analyzing,lancheros2025data}. These issues are especially relevant for English-Vietnamese biomedical NER, where biomedical terms and span boundaries may change across languages.

Translation-based methods provide another way to reduce annotation cost. \citet{Magnini2025} demonstrated the feasibility of LLM-assisted multilingual corpus expansion through a pipeline that combines translation, annotation transfer, format conversion, and manual correction. Building on this direction, En-ViMedNER extends the idea to biomedical NER by preserving ontology-based semantic types and performing entity-level span alignment between English annotations and expert-post-edited Vietnamese text. Moreover, we complement scalable projection with a systematic human audit of projected annotations to quantify projection reliability. Prior work has projected entity labels through machine translation or aligned spans in translated clinical data \citep{jain2019entity,schafer2022cross}. Two common projection strategies are marker-based translation and token-level alignment. EasyProject inserts entity markers into the source before translation \citep{chen-etal-2023-frustratingly}, while alignment tools such as AWESoME learn token correspondences from parallel corpora \citep{dou-neubig-2021-word}. These approaches are less suited to our setting. The former requires markers in the English source before translation, which is incompatible with our workflow, where labels are projected onto the already-finalized, expert-post-edited Vietnamese translation. The latter is weaker because the post-edited text often contains non-literal paraphrases (e.g., "animals" is expanded to "động vật thí nghiệm" meaning "laboratory animals"). Other studies compare direct transfer with translation-based NER, showing that translation can be useful but requires careful preservation of entity spans and labels \citep{gaschi2023multilingual}. More recently, cross-lingual NER has also been formulated as a generation task, where models produce the target-language sentence together with explicit entity tags \citep{yang2022crop}.

Building on these directions, we construct En-ViMedNER as a parallel English-Vietnamese biomedical NER corpus and evaluate it in two complementary settings. First, we examine whether English biomedical supervision can transfer to Vietnamese NER through multilingual encoder fine-tuning. Second, we study cross-lingual NER, comparing prompt-based LLM generation with fine-tuned encoder-decoder models under both direct and cascaded translation-and-NER pipelines.

\section{Corpus Construction}

\begin{figure}
   \centering
   \includegraphics[width=\columnwidth]{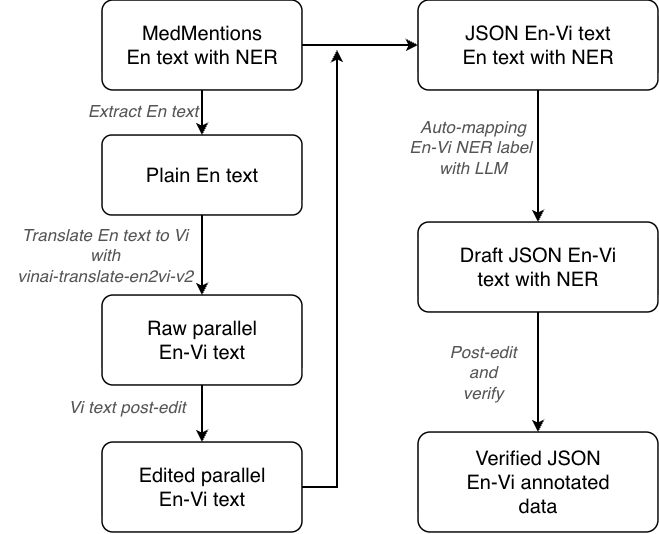}
   \caption{Pipeline for constructing English-Vietnamese biomedical NER data from MedMentions.}
   \label{fig:corpus-creation-diagram}
\end{figure}

We construct \textbf{En-ViMedNER} from MM-ST21pv using three main steps: translation and post-editing, label projection, and human verification, as shown in Figure~\ref{fig:corpus-creation-diagram}.

\subsection{Machine Translation and Medical Expert Post-editing}

We extract English sentences from MM-ST21pv and remove overlapping or unlabeled sentences, resulting in 44,892 sentences. These sentences are translated into Vietnamese using \texttt{vinai-translate-en2vi-v2}~\cite{vinaitranslate}. {All post-editors were Vietnamese medical experts, including final-year medical students and medical doctors, who reviewed translations for terminology accuracy, fluency, and preservation of the original biomedical meaning. During post-editing, they corrected mistranslations and unnatural phrasing while following medical terminology references and English–Vietnamese medical dictionary\footnote{\url{https://meddict-vinuni.com/}} when deciding whether a term should be translated or retained in English. Medical abbreviations and English technical terms were retained when this was the conventional Vietnamese usage; otherwise, they were translated using the medically appropriate Vietnamese equivalent.

\subsection{LLM-assisted Label Projection}

We then convert the parallel sentences into JSON templates compatible with Label Studio. Each template keeps the original English entity spans and labels, while the Vietnamese annotation fields are left blank for label projection (Appendix~\ref{sec:appendix JSON data format}). We use \texttt{gpt-5.2}\footnote{With \texttt{none} reasoning effort, medium verbosity, and top-p \texttt{0.98}. We choose this model because, in a preliminary test on one abstract (98 labels), it achieves 100\% exact match accuracy, compared with 91.8\% for \texttt{gemini-3-pro-preview}~\citep{google2026gemini30preview}.}~\citep{openai2026gpt52} to map each English entity to its corresponding Vietnamese span and assign the same semantic label.


After projection, we apply automatic consistency checks between the English source entities and the projected Vietnamese annotations (Appendix \ref{sec:projection_checks}). Each English entity is linked to its projected Vietnamese counterpart through a shared identifier, enabling entity-level comparison. A sentence pair is flagged when an expected Vietnamese entity is missing, the projected character offsets do not match the extracted Vietnamese text, projected entities overlap unexpectedly, duplicate annotations occur, or an entity begins or ends with suspicious modifiers. Flagged sentence pairs are added to a review together with the detected inconsistency and are manually checked against the English source sentence and Vietnamese translation on Label Studio interface (Appendix \ref{sec:appendix Label studio}). Confirmed errors, including untranslated entities, omitted abbreviations, and incorrect entity boundaries, are then corrected. At the corpus scale, these automatic checks identify approximately 1.77\% of sentence pairs as requiring manual review.
\begin{table}[t]
\centering
\small
\resizebox{1\columnwidth}{!}{%
\begin{tabular}{lrrr}
\hline
\textbf{Split} &
\textbf{\shortstack{Abstract\\pairs}} &
\textbf{\shortstack{Sentence\\pairs}} &
\textbf{\shortstack{Aligned entity-mention\\pairs}} \\
\hline
train & 2,635 & 27,013 & 122,017 \\
dev  & 878  & 8,932  & 40,844  \\
test  & 879  & 8,947  & 40,088  \\
mini-test* & -- & 450 & 2,085 \\
\hline
Total & 4,392 & 44,892 & 202,949 \\
\hline
\end{tabular}%
}
\caption{Statistics of the En-ViMedNER dataset across train, development, and test splits. \textbf{Note}: The mini-test set is randomly sampled from the test set for human evaluation and is not counted as an additional split.}
\label{tab:dataset_statistics}
\end{table}

We convert the verified entity spans into BIO-tagged data for the experiments. Following the original MedMentions split, we divide the corpus into \textbf{train}, \textbf{dev}, and \textbf{test} sets with an approximate 60/20/20 ratio, and sample 450 test sentences as a \textbf{mini-test} set for human evaluation. Table~\ref{tab:dataset_statistics} reports the final corpus statistics.

\subsection{Human Verification and Adjudication}

We assess the pre-correction quality of the automatically projected Vietnamese annotations on a randomly sampled mini-test subset of 450 sentences (2,085 entity pairs). Random sampling avoids intentionally selecting either easier or more challenging instances and provides an unbiased estimate of aggregate projection quality. They correspond to approximately 5\% of the full test set, making it a meaningful audit sample for estimating aggregate projection quality. However, random sampling does not ensure sufficient coverage of every rare semantic type, entity-length category, abbreviation, or difficult translation phenomenon. Therefore, the audit results should be interpreted as evidence of overall projection quality rather than as a reliability guarantee for every individual subgroup.

Before manual correction, two authors independently reviewed the projected Vietnamese spans and labels against the English source annotations and the post-edited Vietnamese translations. They checked whether each projected span matched the intended biomedical entity and whether its character boundaries were correct. Projection quality was evaluated using four criteria: text match, start index, end index, and exact match. (Appendix~\ref{sec:appendix projection quality}).

Inter-annotator agreement before discussion was measured using Cohen's $\kappa$ \cite{cohen1960coefficient} at the sentence level. For each of the 450 sentences, annotators judged whether an error occurred under each projection criterion. The overall pooled $\kappa$ was 0.83, computed over sentence-level judgments, indicating substantial to almost-perfect agreement according to the Landis and Koch scale~\cite{landis1977measurement}. 
{Criterion-specific agreement scores are reported in Table~\ref{tab:kappa_agreement}.



After an independent review, annotators and authors discussed all disagreements and suspected error cases. Confirmed projection errors were resolved by consensus and manually corrected in the final mini-test annotations. Importantly, the reported 98.32\% exact-match accuracy is computed before any manual correction, and therefore measures the quality of the original GPT-projected annotations rather than the final gold annotations. Thus, the audited mini-test is manually verified and consensus-corrected, whereas the 98.32\% figure quantifies pre-correction projection quality.

\section{Experiments}
\subsection{Experimental Settings}
\label{subsec:experimental-tracks}

We organize our experiments into two main tasks:

\paragraph{Task 1 - Vietnamese NER.} This task tests whether En-ViMedNER provides useful Vietnamese supervision beyond English-supervised transfer and prompt-based LLMs. We compare fine-tuned encoder models (Track 1.1) and prompt-based LLMs (Track 1.2).

\textbf{Track 1.1 - Fine-tuned encoder models}: We evaluate fine-tuned encoder models under two settings. First, in the Vietnamese-supervised setting, PhoBERT, XLM-R, ViHealthBERT, and ViPubMedDeBERTa are fine-tuned on the Vietnamese training split and evaluated on the Vietnamese test split. Second, in the English-supervised setting, multilingual XLM-R models are fine-tuned on the English training split and evaluated directly on the same Vietnamese test split. Results are reported on the full test set, with mini-test scores included for comparison with prompt-based LLMs in Track 1.2.

\textbf{Track 1.2 - Prompt-based LLMs}: We evaluate prompt-based LLMs on the mini-test set. We include both open-source LLMs, such as Qwen and SEA-LION, and closed-source frontier LLMs, such as GPT and Gemini.  In this setting, each model receives a Vietnamese sentence and is asked to reproduce the sentence with BIO biomedical NER tags. We evaluate zero-shot, 1-shot, and 3-shot prompting, where few-shot examples are retrieved from the Vietnamese labeled training pool using embedding similarity.

Overall, these experiments test whether supervised fine-tuning on En-ViMedNER is beneficial for Vietnamese biomedical NER, compared with prompt-based LLMs.

\paragraph{Task 2 - Cross-lingual NER:  English-input/Vietnamese-output.} This task evaluates English-to-Vietnamese cross-lingual NER, where models generate a Vietnamese translation with biomedical entity tags from a plain English sentence. We compare fine-tuned encoder-decoder models (Track 2.1) and prompt-based LLMs (Track 2.2) on the mini-test set.

\textbf{Track 2.1 - Fine-tuned multilingual encoder-decoder models}: We fine-tune umT5-base and NLLB-200-distilled-600M on the parallel data with inline NER tags.
We compare direct end-to-end cross-lingual NER with cascaded pipelines that separate translation and NER into sequential steps. We define three settings:

\begin{itemize}[leftmargin=*, itemsep=1pt, topsep=2pt, parsep=0pt, partopsep=0pt]
    \item \textbf{M1: Direct Trans.+NER}: direct end-to-end fine-tuning, where the input is a plain English sentence and the output is a tagged Vietnamese sentence.

    \item \textbf{M2: Trans.$\rightarrow$NER}: a cascaded pipeline where the \textbf{Trans. only} component first translates the plain English input into a plain Vietnamese sentence, and the \textbf{NER only} component then tags the translated Vietnamese sentence.

    \item \textbf{M3: NER$\rightarrow$Trans.}: a cascaded pipeline where the \textbf{NER only} component first tags the plain English input, and \textbf{Tag-aware Trans.} component then translates the tagged English sentence into a tagged Vietnamese sentence.
\end{itemize}
An overview diagram (Figure \ref{fig:track2.1_translation_ner_setup}) and details of \textbf{Trans. only/NER only/Tag-aware Trans.} components are provided in Appendix~\ref{sec:appendix Track 2.1}.

\textbf{Track 2.2 - Prompt-based LLMs}: We evaluate prompt-based LLMs under the same English-input/Vietnamese-output setting. We evaluate both open-source and closed-source LLMs under zero-shot, 1-shot, and 3-shot prompting, where few-shot examples are retrieved from the parallel labeled training pool using embedding similarity.

Overall, these experiments test whether supervised fine-tuning on En-ViMedNER is beneficial for cross-lingual NER, compared with prompt-based LLMs.

\subsection{Models}
\label{subsec:models}
We evaluate six distinct groups of models (see Appendix~\ref{sec:appendix Implementation Details} for details of the model setups, training configurations, hyperparameters, and LLM's prompt templates):
\vspace{-0.2em}
\begin{itemize}[leftmargin=*, itemsep=1pt, topsep=2pt, parsep=0pt, partopsep=0pt]
    \item \textbf{Monolingual encoders.} We use \texttt{PhoBERT-base/base-v2/large}, which are RoBERTa-style models pre-trained for Vietnamese \citep{nguyen-tuan-nguyen-2020-phobert}.
    \item \textbf{Vietnamese biomedical and healthcare encoders.} We use \texttt{ViHealthBERT-word/syllable}, domain-specific pre-trained language models for Vietnamese healthcare text mining \citep{nguyen2022vihealthbert}; \texttt{ViPubMedDeBERTa-xsmall/base}, which are DeBERTa-style models pre-trained for Vietnamese biomedical text \citep{tien2023vipubmeddeberta}.
    \item \textbf{Multilingual encoders.} We use \texttt{XLM-R-base/large}, multilingual masked language models pre-trained on 100 languages \citep{conneau-etal-2020-unsupervised}.
     \item \textbf{Multilingual encoder-decoder models.} We use \texttt{umT5-base}, a multilingual T5-style encoder-decoder model \citep{umt5-chung-etal-2023-unimax}, and \texttt{NLLB-200-distilled-600M}, a distilled model from the NLLB-200 translation family \citep{nllb-team-2022-no}. These models provide a controlled comparison between NLLB-200-distilled-600M as a translation-specialized multilingual model, and umT5-base (580M) as a general multilingual text-to-text model. We exclude larger umT5 variants and mBART from this comparison because their substantially different capacities would introduce additional confounding factors.
    \item \textbf{Open-source LLMs.} We use \texttt{Qwen2.5-3B/7B-Instruct} \citep{qwen25}, and \texttt{Qwen-SEA-LION-v4-8B-VL} a Southeast Asia-oriented model adapted from Qwen \citep{aisingapore2025qwensealionv4}. These models provide a controlled comparison of model scale and regional adaptation within the same model family. Specifically, Qwen2.5 enables comparison across parameter scales (3B/7B), while SEA-LION allows us to examine the effect of Southeast Asia-oriented adaptation for Vietnamese biomedical NER prompting.
    \item \textbf{Closed-source LLMs.} We use \texttt{gpt-5.4} and \texttt{gemini-3.1-flash-lite-preview} as frontier closed-source LLMs through their official APIs \citep{openai2026gpt54,google2026gemini31flashlite}.
\end{itemize}

\subsection{Evaluation Metrics}

For Task~1, we report entity-level precision, recall, and F1 using \texttt{seqeval} \citep{seqeval}. For Task~2, we evaluate both translation quality and biomedical NER performance. We use \texttt{SacreBLEU} \citep{Post2018ACF} to measure the generated Vietnamese sentence against the reference Vietnamese translation. For NER performance, we report two metrics: \textbf{Exact-F1} and \textbf{Judge-F1}. Exact-F1 is computed with \texttt{seqeval} after converting inline entity tags into BIO sequences, requiring exact surface-form and label matching. Judge-F1 is computed from the LLM-as-a-judge protocol described in the next subsection.

\paragraph{LLM-as-a-judge Evaluation.}
For the evaluation of Track 2.1 and Track 2.2, exact span matching is often too restrictive. 
In these settings, the input sentence is written in English, while the model output is a Vietnamese sentence with NER tags. As a result, the predicted entity span may be a valid translation or paraphrase of the gold Vietnamese entity, rather than an exact text match. 
This limits direct F1 computation against the gold Vietnamese annotation, especially when multiple but semantically equivalent translations are possible.

To address this issue, we introduce an LLM-as-a-judge evaluation protocol that applies to experiments where the input and output languages differ. We use \texttt{gpt-5.4-mini} with zero-shot prompting as the judge \citep{openai2026gpt54mini}. Given a model prediction and its extracted entity list, together with the gold sentence and gold entity list, the judge is asked to match each gold entity with the corresponding predicted entity, if any. For each gold entity, the judge evaluates whether the predicted text span and entity label are correct. Each case is categorized as \textit{correct}, \textit{incorrect}, or \textit{miss}. The judge is also asked to identify extra predicted entities that do not correspond to any gold entity (see judging prompt at Appendix~\ref{sec:appendix Prompt templates}).
We finally convert the judge decisions into entity-level F1 counts as follows:
\begin{itemize}[leftmargin=*, itemsep=1pt, topsep=2pt, parsep=0pt, partopsep=0pt]
    \item Exact span and tag match: counted as one TP.
    \item Any span or tag mismatch, including cases where both are incorrect, is counted as one FP for the prediction and one FN for the corresponding gold entity.
    \item Missed gold entity: counted as one FN.
    \item Extra predicted entity: counted as one FP.
    \item Invalid LLM outputs (reported as \textbf{Inv.}) are not discarded from evaluation. An output is marked as invalid if it cannot be parsed into the required JSON/tagged format or if it contains no valid entity tags (i.e., all tokens are tagged as ``O''). For such cases, the prediction set is treated as empty: all gold entities are counted as FN, and no predicted entities are counted as TP.
\end{itemize}

To verify the reliability of this protocol, we measure human-LLM agreement on 100 randomly sampled outputs from the NLLB-M1 setting. Human annotators apply the same criteria as the LLM judge, covering entity-span matching, label matching, missed entities, and extra predictions, and we report Cohen's $\kappa$ to assess the consistency of the LLM judge with human judgment.

\section{Results}
\subsection{Task 1 - Vietnamese NER}

\begin{table*}[!t]
\centering
\scriptsize
\setlength{\tabcolsep}{3.0pt}
\renewcommand{\arraystretch}{1.05}
\resizebox{0.6\textwidth}{!}{%
\begin{tabular}{lcccccc}
\hline
\textbf{Model}
& \multicolumn{3}{c}{\textbf{mini-test set}}
& \multicolumn{3}{c}{\textbf{Full test set}} \\
\cline{2-4}
\cline{5-7}
& \textbf{Precision} & \textbf{Recall} & \textbf{F1}
& \textbf{Precision} & \textbf{Recall} & \textbf{F1} \\
\hline

\multicolumn{7}{l}{\textbf{Vietnamese-supervised setting}} \\
PhoBERT-base & 54.28 & 51.83 & 53.03 & 52.65 & 49.98 & 51.28 \\
PhoBERT-base-v2 & 53.72 & 51.68 & 52.68 & 52.60 & 50.42 & 51.49 \\
PhoBERT-large & 53.41 & 51.63 & 52.51 & 52.19 & 50.84 & 51.50 \\
XLM-R-base & 53.53 & 52.12 & 52.81 & 50.68 & 48.92 & 49.78 \\
XLM-R-large & 54.91 & 51.63 & 53.22 & 52.66 & 50.29 & 51.45 \\
ViHealthBERT-word & 53.70 & 51.25 & 52.45 & 52.72 & 50.51 & 51.59 \\
ViHealthBERT-syllable & 52.37 & 50.00 & 51.16 & 50.42 & 48.47 & 49.42 \\
ViPubMedDeBERTa-xsmall & 54.00 & 53.56 & \textbf{53.78} & 52.35 & 52.74 & 52.54 \\
ViPubMedDeBERTa-base & 54.03 & 53.22 & 53.62 & 52.86 & 52.54 & \textbf{52.70} \\

\hline
\multicolumn{7}{l}{\textbf{English-supervised setting}} \\
XLM-R-base & 30.26 & 37.72 & 33.58 & 27.80 & 35.48 & 31.12 \\
XLM-R-large & 35.14 & 41.26 & 37.95 & 33.54 & 40.13 & 36.54 \\

\hline
\multicolumn{7}{l}{\textbf{Open-source LLMs}}\\
Qwen2.5-3B-Instruct-0shot & 4.03 & 0.83 & \multicolumn{1}{c}{1.37} & \multicolumn{3}{c}{} \\
Qwen2.5-3B-Instruct-1shot & 8.08 & 2.09 & \multicolumn{1}{c}{3.32} & \multicolumn{3}{c}{} \\
Qwen2.5-3B-Instruct-3shot & 9.76 & 2.14 & \multicolumn{1}{c}{3.50} & \multicolumn{3}{c}{} \\
Qwen2.5-7B-Instruct-0shot & 2.14 & 0.24 & \multicolumn{1}{c}{0.44} & \multicolumn{3}{c}{} \\
Qwen2.5-7B-Instruct-1shot & 9.57 & 2.62 & \multicolumn{1}{c}{4.12} & \multicolumn{3}{c}{} \\
Qwen2.5-7B-Instruct-3shot & 12.25 & 2.72 & \multicolumn{1}{c}{4.45} & \multicolumn{3}{c}{} \\
Qwen-SEA-LION-v4-8B-VL-0shot & 8.61 & 0.87 & \multicolumn{1}{c}{1.59} & \multicolumn{3}{c}{} \\
Qwen-SEA-LION-v4-8B-VL-1shot & 16.10 & 6.65 & \multicolumn{1}{c}{9.41} & \multicolumn{3}{c}{} \\
Qwen-SEA-LION-v4-8B-VL-3shot & 21.22 & 9.47 & \multicolumn{1}{c}{13.09} & \multicolumn{3}{c}{} \\

\cline{1-4}
\multicolumn{7}{l}{\textbf{Closed-source LLMs}}\\
gpt-5.4-0shot & 11.79 & 12.99 & \multicolumn{1}{c}{12.36} & \multicolumn{3}{c}{} \\
gpt-5.4-1shot & 20.04 & 22.50 & \multicolumn{1}{c}{21.20} & \multicolumn{3}{c}{} \\
gpt-5.4-3shot & 32.19 & 36.26 & \multicolumn{1}{c}{34.11} & \multicolumn{3}{c}{} \\
gemini-3.1-flash-lite-preview-0shot & 13.16 & 12.26 & \multicolumn{1}{c}{12.70} & \multicolumn{3}{c}{} \\
gemini-3.1-flash-lite-preview-1shot & 23.55 & 24.92 & \multicolumn{1}{c}{24.21} & \multicolumn{3}{c}{} \\
gemini-3.1-flash-lite-preview-3shot & 35.10 & 37.95 & \multicolumn{1}{c}{36.47} & \multicolumn{3}{c}{} \\

\cline{1-4}
\end{tabular}%
}
\caption{Performance comparison of encoder fine-tuning and prompt-based LLMs on Vietnamese biomedical NER. The input is a plain-text Vietnamese sentence, and the output is Vietnamese NER annotations. Bold numbers indicate the highest F1 score within each evaluation set.}
\label{tab:vi-vi}
\end{table*}

Table~\ref{tab:vi-vi} compares three settings for Vietnamese NER: Vietnamese-supervised encoder fine-tuning, English-supervised multilingual encoder fine-tuning, and prompt-based LLMs.

\paragraph{Vietnamese supervision provides a clear advantage.}
Vietnamese-supervised encoder models consistently outperform English-supervised multilingual encoder models on both evaluation sets. 
On the full test set, \texttt{ViPubMedDeBERTa-base} achieves 52.70 F1, outperforming the best English-supervised model, \texttt{XLM-R-large}, by 16.16 points. 
The same pattern holds on the mini-test set, where \texttt{ViPubMedDeBERTa-xsmall} obtains 53.78 F1 compared with 37.95 F1 for \texttt{XLM-R-large}. 
These results indicate that multilingual pretraining and English supervision cannot replace target-language annotations, highlighting the value of En-ViMedNER for Vietnamese biomedical entity boundaries, terminology, and labels.

\paragraph{Prompting improves with examples but remains below fine-tuning.}
Few-shot prompting improves both open-source and closed-source LLMs, but the gains remain insufficient to match supervised encoder fine-tuning. For example, \texttt{gemini-3.1-flash-lite-preview} improves from 12.70 F1 in the \texttt{zero-shot} setting to 36.47 F1 with \texttt{3-shot} prompting. Among open-source LLMs, \texttt{Qwen-SEA-LION-v4-8B-VL} improves from 1.59 to 13.09 F1. Nevertheless, the best prompted model, \texttt{gemini-3.1-flash-lite-preview-3shot}, remains 17.31 F1 points behind the best Vietnamese-supervised encoder. These results suggest that in-context examples help LLMs follow the BIO tagging format, but do not replace task-specific Vietnamese biomedical supervision.

Overall, the findings from Task 1 demonstrate that En‑ViMedNER constitutes a valuable resource for developing robust Vietnamese biomedical named entity recognition (NER) systems, rather than serving solely as an evaluation benchmark.

\subsection{Task 2 - Cross-lingual NER}

Table~\ref{tab:translation_ner_llm_results} compares fine-tuned encoder-decoder models and prompt-based LLMs on the mini-test set. We report entity-level precision, recall, and Judge-F1 using the LLM-as-a-judge protocol, together with Exact-F1 based on exact entity matching against the gold annotations. The reliability of the judging protocol is further supported by human-LLM agreement, with Cohen's $\kappa$ = 0.87.

\paragraph{Fine-tuned encoder-decoder models outperform prompting.}
Fine-tuning on En-ViMedNER parallel data gives the strongest overall performance. 
Under Judge-F1, the best fine-tuned setting is NLLB-M1, which reaches 45.44 F1 and outperforms the best closed-source LLMs, including \texttt{gpt-5.4-3shot} and \texttt{gemini-3.1-flash-lite-preview-3shot}, by 2.80 and 3.89 points, respectively. 
The same trend holds under the stricter Exact-F1 metric: NLLB-M3 achieves the highest Exact-F1 score of 15.98, above the best closed-source LLM result of 10.40. 
These results suggest that paired English inputs and Vietnamese tagged outputs provide useful supervision for learning both translation and biomedical entity transfer.

\paragraph{Neither end-to-end nor cascaded pipelines are uniformly superior.} 
Under Judge-F1, direct end-to-end generation performs best for both models: umT5-M1 achieves 26.88 F1, while NLLB-M1 achieves 45.44 F1. However, under Exact-F1, the \textbf{NER$\rightarrow$Trans.} pipeline performs best for both models, reaching 7.33 for umT5-M3 and 15.98 for NLLB-M3. This suggests that direct generation is more effective when semantically equivalent entity translations are accepted, whereas the tag-first cascaded pipeline may better preserve exact entity forms. Therefore, the relative advantage of end-to-end and cascaded pipelines depends on the evaluation criterion.

\begin{table}[!t]
\centering
\scriptsize
\setlength{\tabcolsep}{2.0pt}
\renewcommand{\arraystretch}{1.05}
\resizebox{\linewidth}{!}{%
\begin{tabular}{lcccccc}
\hline
\multirow{2}{*}{\textbf{Setting/Model}}
& \multicolumn{6}{c}{\textbf{Metrics}} \\
\cline{2-7}
& \textbf{BLEU} 
& \textbf{Prec.} 
& \textbf{Rec.} 
& \textbf{Judge-F1} 
& \textbf{Exact-F1} 
& \textbf{Inv.} \\
\hline
\multicolumn{7}{l}{\textbf{Fine-tuned multilingual encoder-decoder models}}\\
umT5-M1: Direct Trans.+NER 
& 29.85 & 24.41 & 29.90 & 26.88 & 6.01 & 0 \\
umT5-M2: Trans.$\rightarrow$NER 
& 35.98 & 32.80 & 19.71 & 24.62 & 5.17 & 0 \\
umT5-M3: NER$\rightarrow$Trans. 
& 31.15 & 30.64 & 23.56 & 26.64 & 7.33 & 0 \\

\cline{2-7}

NLLB-M1: Direct Trans.+NER 
& 48.63 & 44.31 & 46.63 & \textbf{45.44} & 15.77 & 0 \\
NLLB-M2: Trans.$\rightarrow$NER 
& 51.83 & 43.77 & 38.33 & 40.87 & 13.64 & 0 \\
NLLB-M3: NER$\rightarrow$Trans. 
& 46.90 & 41.32 & 36.95 & 39.02 & \textbf{15.98} & 0 \\

\hline
\multicolumn{7}{l}{\textbf{Open-source LLMs}} \\
Qwen2.5-3B-Instruct-0shot 
& 35.81 & 8.73 & 2.14 & 3.43 & 0.91 & 114 \\
Qwen2.5-3B-Instruct-1shot 
& 36.39 & 8.98 & 2.77 & 4.23 & 1.80 & 178 \\
Qwen2.5-3B-Instruct-3shot 
& 38.73 & 11.86 & 4.17 & 6.18 & 3.33 & 158 \\
Qwen2.5-7B-Instruct-0shot 
& 40.79 & 7.93 & 1.26 & 2.18 & 1.99 & 194 \\
Qwen2.5-7B-Instruct-1shot 
& 36.71 & 15.20 & 3.79 & 6.06 & 3.92 & 201 \\
Qwen2.5-7B-Instruct-3shot 
& 43.66 & 22.28 & 8.16 & 11.94 & 5.34 & 115 \\
Qwen-SEA-LION-v4-8B-VL-0shot 
& 50.06 & 21.00 & 3.88 & 6.55 & 6.17 & 292 \\
Qwen-SEA-LION-v4-8B-VL-1shot 
& 49.99 & 22.59 & 8.06 & 11.88 & 9.19 & 250 \\
Qwen-SEA-LION-v4-8B-VL-3shot 
& 52.89 & 30.23 & 13.25 & 18.43 & 9.85 & 228 \\

\cline{1-7}
\multicolumn{7}{l}{\textbf{Closed-source LLMs}} \\
gpt-5.4-0shot 
& 56.59 & 28.94 & 35.68 & 31.96 & 4.43 & 0 \\
gpt-5.4-1shot 
& \textbf{56.60} & 33.24 & 41.48 & 36.91 & 7.33 & 0 \\
gpt-5.4-3shot 
& 53.34 & 38.74 & 47.42 & 42.64 & 9.84 & 0 \\
gemini-3.1-flash-lite-preview-0shot 
& 55.03 & 30.03 & 34.14 & 31.95 & 4.43 & 0 \\
gemini-3.1-flash-lite-preview-1shot 
& 52.18 & 34.70 & 42.39 & 38.17 & 8.05 & 0 \\
gemini-3.1-flash-lite-preview-3shot 
& 48.80 & 38.59 & 45.00 & 41.55 & 10.40 & 0 \\

\hline
\end{tabular}
}
\caption{Performance comparison across encoder-decoder fine-tuning and prompt-based LLMs on the mini-test set. The input is a plain-text English sentence, and the output is Vietnamese NER annotations. \textbf{Inv.} denotes the number of generated sentences without valid entity tags. Bold numbers indicate the highest value in the corresponding metric column.}
\label{tab:translation_ner_llm_results}
\end{table}

\paragraph{Open-source LLMs struggle with both tagging quality and output validity.}
Open-source LLMs remain substantially weaker than fine-tuned encoder-decoder models and closed-source LLMs under Judge-F1. The best open-source result is \texttt{Qwen-SEA-LION-v4-8B-VL-3shot}, which reaches 18.43 Judge-F1 and 9.85 Exact-F1, still far below the best fine-tuned encoder-decoder results (NLLB-M1). A major issue is invalid formatting: open-source LLMs often fail to produce valid Vietnamese NER-tagged outputs, with invalid counts ranging from 114 to 292 out of 450 examples. This suggests that these models often fail to consistently follow the required biomedical tagging format in the cross-lingual generation setting.

\paragraph{Higher BLEU does not always imply better NER.}
\begin{figure}[!t]
    \centering
    \includegraphics[width=1\linewidth]{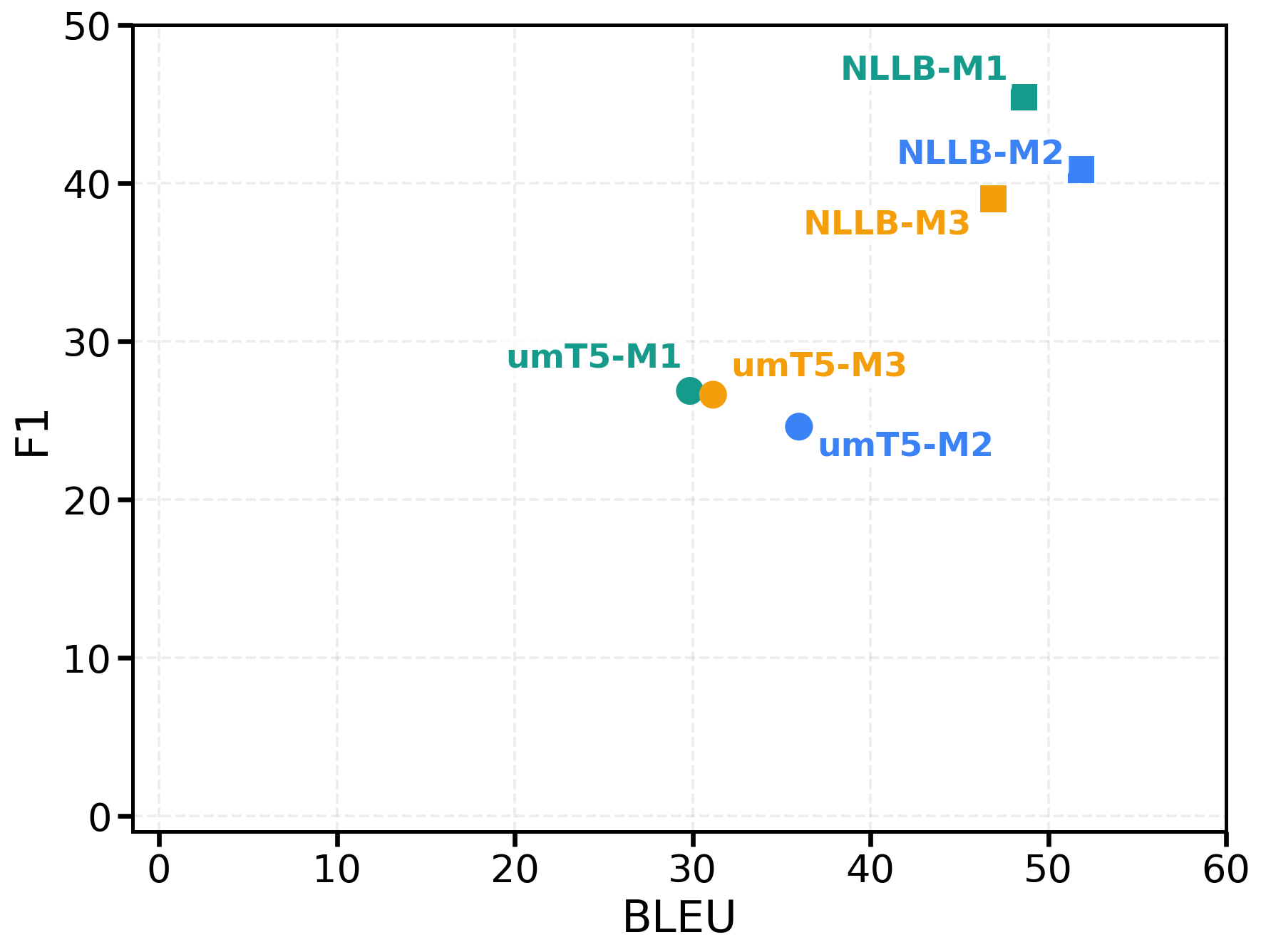}
    \caption{Relationship between translation quality and NER performance for encoder-decoder cross-lingual NER on the mini-test set. Circles indicate \texttt{umT5-base}, squares indicate \texttt{NLLB-200-distilled-600M}, and each point represents one experimental setting. \textit{Note}: Colors are used only as a secondary visual cue to group the same setting across models; the figure remains interpretable through marker shapes and direct labels.}
\label{fig:envi_bleu_f1_comparison}
\end{figure}

Figure~\ref{fig:envi_bleu_f1_comparison} shows that translation quality and NER performance are not perfectly aligned. For example, NLLB-M2 obtains the highest BLEU score among the main NLLB systems, but its F1 is lower than NLLB-M1. A similar pattern appears for umT5: M2 has the highest BLEU score, but M1 gives the best F1. This indicates that good sentence-level translation is not enough for biomedical NER generation. The model must also preserve entity boundaries and labels during generation.

Overall, the findings from Task 2 demonstrate that En-ViMedNER's parallel English-Vietnamese annotations provide a valuable foundation for cross-lingual biomedical NER, enabling models to jointly learn translation, entity-boundary preservation, and biomedical label transfer.

\section{Conclusion and Future Work}
We presented En-ViMedNER, the first parallel English-Vietnamese biomedical NER corpus with UMLS semantic annotations, comprising 4,392 PubMed abstracts and 202,949 aligned entity-mention pairs across 21 UMLS semantic types. The corpus is constructed via a hybrid pipeline of machine translation, expert post-editing, LLM-assisted projection, and human verification (Cohen's $\kappa$ = 0.83, entity-level exact match = 98.32\%). 

Our experiments demonstrate the value of En-ViMedNER in two settings. For Vietnamese biomedical NER, Vietnamese-supervised encoder models consistently outperform both English-supervised multilingual encoders and prompt-based LLMs, showing that multilingual pretraining and prompting cannot replace target-language biomedical supervision. For cross-lingual NER, fine-tuned encoder-decoder models outperform prompt-based LLMs, indicating that the parallel annotations provide useful supervision for jointly learning translation, entity-boundary preservation, and biomedical label transfer. The results further show that translation quality alone is insufficient for cross-lingual biomedical NER, and that end-to-end and cascaded pipelines offer different advantages depending on the evaluation criterion.

Future work will explore extensions to other low-resource Southeast Asian languages, particularly Indonesian and Thai, where emerging medical NER resources provide useful starting points for evaluating the portability of our construction pipeline. We will also investigate additional experimental settings involving open-source LLM fine-tuning, LLM-as-a-judge reliability \citep{wang2025trustjudge}, and cascaded medical translation-NER pipelines.

\section*{Limitations}
Despite our efforts to ensure corpus quality through expert post-editing of translations, as well as review and correction after LLM-assisted label projection, there are limitations that users of En-ViMedNER should be aware of.
\begin{itemize}[leftmargin=*, itemsep=1pt, topsep=2pt, parsep=0pt, partopsep=0pt]
   \item \textbf{Sentence-level vs. abstract-level context}: Some Vietnamese translations only make sense in a broad context of an abstract. For example, the translation \textit{animals --> động vật thí nghiệm} assumes a laboratory-experiment context, which is available at the abstract level but not at the sentence level. Because our annotation and evaluation operate at the sentence level, the meaning of some words may appear less natural than they would in full context.
   \item \textbf{Surface-form ambiguity}: There are cases when the same surface form appears multiple times in a sentence, some are translated but some are kept as the original text. For example, \textit{HCC} appears twice in sentence \textit{Mep1A is overexpressed in most HCC and induces HCC cell migration and invasion.}, the first \textit{HCC} is kept as-is while the second \textit{HCC} is translated to \textit{ung thư gan}. Cases like this may confuse trained NER models and reduce surface-level learnability. 
   \item \textbf{Domain coverage}: En-ViMedNER is built from PubMed abstracts, which are formal biomedical literature. Therefore, models trained on En-ViMedNER may underperform on less formal text like clinical notes and patient-facing health communication.
   \item \textbf{Downstream use}: Models trained on En-ViMedNER should not be assumed to perform reliably on clinical notes, patient-facing text, or safety-critical medical applications without additional validation.
\end{itemize}

\section*{Ethical considerations}
En-ViMedNER follows the licensing conditions of both MedMentions and UMLS. Since MedMentions is released under CC0, we can redistribute the translated dataset. We only retain UMLS semantic type codes (e.g., \texttt{T047}) as reference labels, and do not publish UMLS concept names, definitions, or hierarchy information. Users who need detailed content must access it through their own UMLS accounts under the official UMLS license.

 The corpus is derived from PubMed abstracts and MedMentions rather than clinical notes or patient records, reducing the risk of personally identifying health information.

\section*{Use of AI tools}
We used Grammarly\footnote{https://app.grammarly.com/} for grammar checking and readability improvement during manuscript preparation. All scientific claims, analyses, and final writing decisions remain the responsibility of the authors.

\section*{Acknowledgments}
This research was undertaken within the framework of the \textbf{\textit{Cross-College project Robust Vietnamese–English Clinical and Educational Medical Translation}} (Project ID: VUNI.2324.CC06), jointly undertaken by the College of Engineering \& Computer Science and the College of Health Sciences at VinUniversity.

Nhu Vo acknowledges financial support from the Vingroup Scholarship and the University of Technology Sydney. Dung D. Le acknowledges partial funding from the Center for AI Research at VinUniversity.

\bibliography{custom}

\appendix
These appendix sections provide additional details that support the main paper. 
\section{JSON data format for label mapping}
\label{sec:appendix JSON data format}

Full-abstract prompting caused ambiguity when the same entity surface form appeared multiple times across sentences or when an entity was translated differently depending on local context. We therefore perform annotation mapping at the sentence level. For each English-Vietnamese sentence pair, the input JSON contains the English sentence, the post-edited Vietnamese translation, and the list of English entity spans with their UMLS semantic type labels. The Vietnamese annotation fields are left blank and filled by the LLM during projection.

This format allows the projection model to focus on local sentence-level alignment while preserving the original English entity labels. It also makes the projected spans easy to inspect in Label Studio, where annotators can verify the Vietnamese entity text, start and end offsets, and semantic type. Figure~\ref{fig:json-example} shows an example of the JSON format used for label mapping.
\begin{figure*}[!t]
   \centering
   \includegraphics[width=\textwidth]{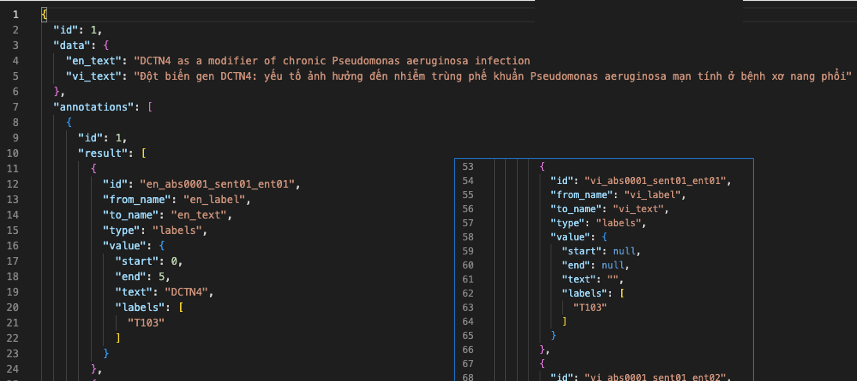}
   \caption{Example of JSON data format for label mapping.}
   \label{fig:json-example}
\end{figure*}

\section{Human verification and adjudication details}
\subsection{Automatic consistency checks and projection review}
\label{sec:projection_checks}
After LLM-assisted label projection, we apply automatic consistency checks to identify
potential inconsistencies between English entities and their projected
Vietnamese counterparts. The checking procedure consists of the following steps:

\begin{enumerate}
    \item \textbf{Match English and Vietnamese entities.}
    Each English entity is linked to its projected Vietnamese entity using
    their shared identifier, such as
    \texttt{en\_abs[xxxx]\_sent[yy]\_ent[zz]} and
    \texttt{vi\_abs[xxxx]\_sent[yy]\_ent[zz]}.

    \item \textbf{Check each sentence pair automatically.}
    A sentence pair is flagged if at least one of the following conditions is
    satisfied:
    \begin{itemize}
        \item An English entity exists, but the corresponding Vietnamese entity
        is missing.
        \item The start-end offsets of a Vietnamese entity do not match the
        actual text span.
        \item Two English entities are separate, but their corresponding
        Vietnamese entities overlap.
        \item An English entity is duplicated with the same label and the same
        span.
        \item A Vietnamese entity begins or ends with suspicious words such as
        ``mỗi'' (``per''), ``này'' (``this/these''), or ``khác'' (``other'').
    \end{itemize}

    \item \textbf{Review queue.}
    Any sentence pair meeting at least one condition is copied to a review
    set, together with the affected entities and the detected inconsistency.

    \item \textbf{Manually verify the flagged pairs.}
    Annotators compare the English source sentence, the Vietnamese translation,
    and their entity spans. This step determines the actual cause, such as an
    untranslated entity, an omitted abbreviation, or an incorrect entity
    boundary.

    \item \textbf{Correct the annotations.}
    The Vietnamese entity text, label, or character offsets are corrected under
    the domain expert's consultation.
\end{enumerate}

\subsection{Manual data checking with Label Studio example}
\label{sec:appendix Label studio}
\begin{figure*}
    \centering
    \includegraphics[width=1\linewidth]{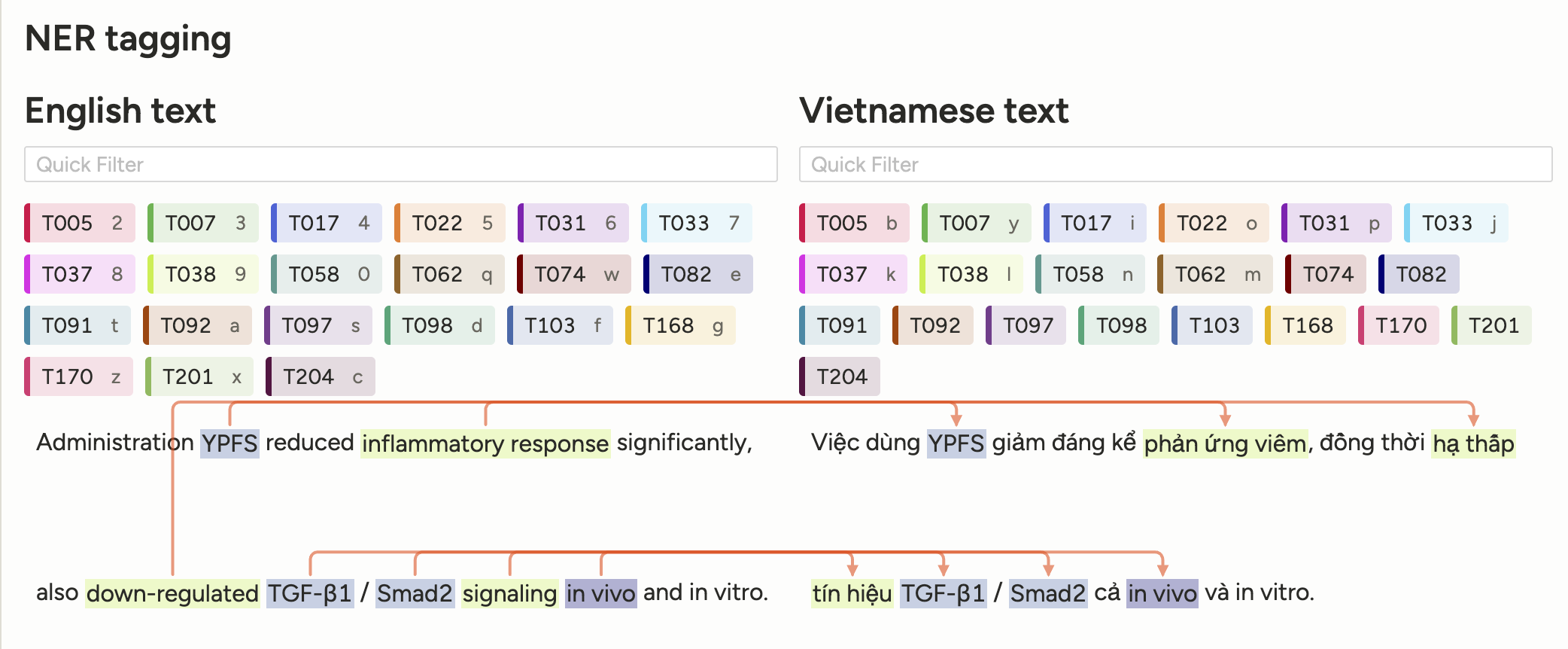}
    \caption{Screenshot of manual data checking in Label Studio.}
    \label{fig:label_studio_example}
\end{figure*}
Figure~\ref{fig:label_studio_example} shows a screenshot of the Label Studio interface used for manual verification. The interface displays each English sentence and its post-edited Vietnamese translation side by side, together with the projected biomedical entity labels. Annotators used this view to check whether the Vietnamese spans correctly matched the English source entities, whether span boundaries were accurate, and whether the UMLS semantic types were preserved. This step helped identify projection errors such as incorrect boundaries, untranslated abbreviations, and inconsistent translations of the same biomedical entity. The label panel also supports toggling the visibility of individual label pairs, helping annotators focus on specific entities during verification.

\subsection{Projection quality evaluation}
\label{sec:appendix projection quality}
Projection quality was evaluated using four criteria: text match, start index, end index, and exact match. Text match checks whether the projected Vietnamese mention matches the intended biomedical entity. Start and end indices assess whether the projected character-level span boundaries are correct. Exact match requires the entity text and both boundaries to be correct.

Inter-annotator agreement was computed at the sentence level. For each of the 450 mini-test sentences, annotators judged whether an error occurred under each projection criterion. The overall pooled $\kappa$ was computed by aggregating all paired ratings across the four criteria.

Based on the consensus audit, entity-level projection accuracy over the 2,085 mentions reached:
\begin{itemize}
\item 98.32\% for text match, 
\item 99.04\% for start index, 
\item 98.27\% for end index, and 
\item 98.32\% for exact match before correction. 
\end{itemize}
The 95\% Wilson confidence interval for exact-match accuracy is 97.67\%-98.79\% \citep{wilson1927probable,brown2001interval}.

\subsection{Inter-annotator agreement details}
\label{sec:appendix kappa agreement}

\begin{table}[!t]
\centering
\small
\resizebox{\columnwidth}{!}{%
\begin{tabular}{lccccc}
\hline
\textbf{Metric} & \textbf{N} & \textbf{Agreement} & $\mathbf{P_o}$ & $\mathbf{P_e}$ & $\boldsymbol{\kappa}$ \\
\hline
Text match accuracy  & 450  & 368/450   & 0.8178 & 0.1234 & 0.7921 \\
Start index accuracy & 450  & 393/450   & 0.8733 & 0.1225 & 0.8557 \\
End index accuracy   & 450  & 406/450   & 0.9022 & 0.1227 & 0.8885 \\
Exact match accuracy & 450  & 367/450   & 0.8156 & 0.1236 & 0.7896 \\
\hline
\textbf{Overall pooled} & \textbf{1800} & \textbf{1534/1800} & \textbf{0.8522} & \textbf{0.1230} & \textbf{0.8315} \\
\hline
\end{tabular}%
}
\caption{Inter-annotator agreement between two annotators measured by Cohen's kappa across four evaluation metrics. $P_o$ denotes observed agreement and $P_e$ denotes expected agreement by chance. The overall pooled result is computed by aggregating all paired ratings across the four metrics.}
\label{tab:kappa_agreement}
\end{table}
Table~\ref{tab:kappa_agreement} shows the agreement between two annotators during the quality check of projected Vietnamese annotations. 
The overall Cohen's $\kappa$ score is 0.8315, which indicates strong agreement. Among the four metrics, end-index accuracy has the highest agreement, while exact-match accuracy is lower because it requires the entity text and both span boundaries to be correct. This suggests that most annotation disagreements come from difficult boundary or text-matching cases rather than random annotation noise.

\section{Error analysis}
\label{sec:appendix Error analysis}
We re-examined the mini-test set and identified 35 confirmed projection errors across 30 of the 450 audited sentences. Table~\ref{tab:error_category} summarizes these errors by category. Most errors were boundary-related (57.1\%), while the remainder involved projection failures, untranslated or omitted terms, abbreviation handling, and terminology inconsistencies. Table~\ref{tab:error_semantic_type} further breaks down the errors by UMLS semantic type, showing that T017 and T058 had the largest numbers of confirmed errors, with six cases each, followed by T082 with five cases. Since boundary errors were the dominant category, Table~\ref{tab:boundary_error_subtype} provides a finer-grained analysis of these cases. Among the 20 boundary errors, 12 (60.0\%) were caused by over-projection, where the projected span included unnecessary words or modifiers, while 8 (40.0\%) were due to under-projection, where part of the intended entity span was omitted.

\begin{table*}[t!]
\centering
\small
\begin{tabular}{llrrp{7cm}}
\toprule
\textbf{Code} & \textbf{Error category} & \textbf{Count} & \textbf{Percentage} & \textbf{Description} \\
\midrule
C1 & Boundary errors & 20 & 57.1\% &
Correct semantic region but missing required words or including extra words in the target span. \\

C2 & Projection errors & 5 & 14.3\% &
Failed to map the entity due to changes in translation structure, order, or span arrangement. \\

C3 & Untranslated or omitted terms & 4 & 11.4\% &
Source terms were untranslated or omitted, leaving no corresponding target entity span. \\

C4 & Abbreviation handling & 2 & 5.7\% &
Abbreviations were added, removed, or handled differently, complicating source-target alignment. \\

C5 & Terminology inconsistencies & 4 & 11.4\% &
The target expression did not fully match the intended meaning of the source term. \\

\midrule
\textbf{Total} & & \textbf{35} & \textbf{100\%} & \\
\bottomrule
\end{tabular}
\caption{Distribution of projection errors by error category.}
\label{tab:error_category}
\end{table*}

\begin{table}[!t]
\centering
\small
\resizebox{\columnwidth}{!}{%
\begin{tabular}{lrrrrrrr}
\hline
\textbf{Semantic type} & \textbf{C1} & \textbf{C2} & \textbf{C3} & \textbf{C4} & \textbf{C5} & \textbf{Total} & \textbf{Percentage} \\
\hline
T017 & 3 & 3 & 0 & 0 & 0 & 6 & 17.1\% \\
T058 & 5 & 0 & 0 & 0 & 1 & 6 & 17.1\% \\
T082 & 3 & 0 & 0 & 1 & 1 & 5 & 14.3\% \\
T170 & 2 & 0 & 0 & 1 & 1 & 4 & 11.4\% \\
T033 & 0 & 1 & 1 & 0 & 1 & 3 & 8.6\% \\
T038 & 2 & 1 & 0 & 0 & 0 & 3 & 8.6\% \\
T098 & 1 & 0 & 2 & 0 & 0 & 3 & 8.6\% \\
T103 & 2 & 0 & 0 & 0 & 0 & 2 & 5.7\% \\
T062 & 2 & 0 & 0 & 0 & 0 & 2 & 5.7\% \\
T204 & 0 & 0 & 1 & 0 & 0 & 1 & 2.9\% \\
\hline
Total & 20 & 5 & 4 & 2 & 4 & 35 & 100\% \\
\hline
\end{tabular}%
}
\caption{Breakdown of projection errors by semantic type. C1-C5 correspond to the error categories defined in Table~\ref{tab:error_category}.}
\label{tab:error_semantic_type}
\end{table}

\begin{table*}[!t]
\centering
\small
\setlength{\tabcolsep}{3pt}
\begin{tabularx}{\textwidth}{@{}>{\raggedright\arraybackslash}p{1.0cm}
>{\raggedright\arraybackslash}p{2.8cm}
rrr
>{\raggedright\arraybackslash}X@{}}
\toprule
\makecell{\textbf{Error}\\\textbf{Code}} &
\makecell{\textbf{Boundary Error}\\\textbf{Subtype}} &
\makecell{\textbf{Number of}\\\textbf{Errors}} &
\makecell{\textbf{Percentage of}\\\textbf{All Errors}} &
\makecell{\textbf{Percentage of}\\\textbf{Boundary Errors}} &
\textbf{Description} \\
\midrule
C1.1 & Over-projection / Over-extension & 12 & 34.3\% & 60.0\% & The projected target span includes additional words or modifiers that do not belong to the source entity. \\

C1.2 & Under-projection / Under-extension & 8 & 22.9\% & 40.0\% & The projected target span captures only part of the corresponding entity and omits one or more required components. \\

\midrule
&
Total boundary errors & 20 & 57.1\% & 100\% &\\
\bottomrule
\end{tabularx}
\caption{Breakdown of boundary errors by subtype.}
\label{tab:boundary_error_subtype}
\end{table*}
}

\section{Details of Track 2.1 experimental settings}
\label{sec:appendix Track 2.1}
\begin{figure*}[!t]
   \centering
   \includegraphics[width=0.8\textwidth]{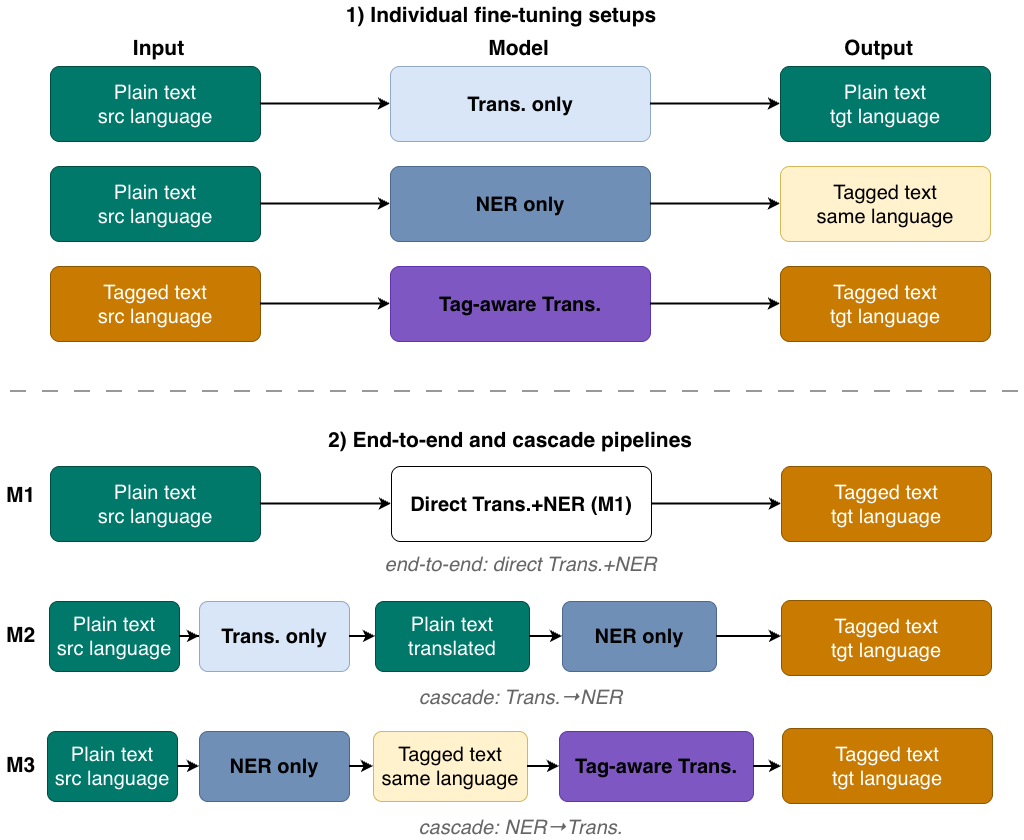}
   \caption{Overview of Track~2.1 settings. The figure compares single-task components with direct and cascaded cross-lingual NER pipelines. \textit{Note}: Colors are used as secondary cues to distinguish plain text, tagged text, and model components; all boxes are directly labeled, so the figure remains readable in grayscale.}
   \label{fig:track2.1_translation_ner_setup}
\end{figure*}

Figure~\ref{fig:track2.1_translation_ner_setup} summarizes the Track~2.1 settings. We separate three component abilities: plain sentence translation, monolingual NER tagging, and tag-preserving translation. 

\textbf{Trans. only} is a translation-only fine-tuned model trained to map a plain English sentence to a plain Vietnamese sentence, without generating NER tags. 

\textbf{NER only} is a monolingual NER fine-tuned model trained to insert entity tags into a plain sentence in the same language. 

\textbf{Tag-aware Trans.} is a tag-aware translation model trained to translate an already tagged English sentence into a tagged Vietnamese sentence while preserving entity labels. 

\textbf{M1} directly generates a tagged Vietnamese sentence from a plain English input, whereas \textbf{M2} and \textbf{M3} represent two cascaded alternatives: translate-then-tag and tag-then-translate.

This setup allows us to compare whether cross-lingual biomedical NER is better handled as a joint generation task or as a pipeline of separately optimized components.

\section{Implementation Details}
\label{sec:appendix Implementation Details}

\paragraph{Encoder fine-tuning.} We fine-tune all parameters of PhoBERT, XLM-R, ViHealthBERT, and ViPubMedDeBERTa for token-level sequence labeling.
We fine-tune the models for \texttt{3} and \texttt{5} epochs with learning rate selected from \texttt{[2e-5, 3e-5, 5e-5]}, batch size \texttt{16}, maximum sequence length \texttt{256}, and warmup ratio \texttt{0.06}, using \texttt{FP16} mixed-precision training.
The best checkpoint is selected by validation \texttt{F1 score}. All experiments use random seed \texttt{42}.

\paragraph{Encoder-decoder fine-tuning.} We formulate the umT5 and NLLB experiments as text-to-text generation tasks with inline entity tags. We fine-tune the models for \texttt{5} epochs with learning rate \texttt{3e-5}, batch size \texttt{16}, gradient accumulation \texttt{2}, maximum source/target length \texttt{256}, warmup ratio \texttt{0.04}, and greedy decoding. Models are evaluated and saved every \texttt{400} steps, with early stopping patience \texttt{3}.

For Trans. only models, the best checkpoint is selected by validation BLEU.

For NER only models, the best checkpoint is selected by validation entity-level F1.

For M1, M2, and M3 models, we select checkpoints by validation BLEU and report NER F1 only in the final evaluation to avoid using judge-based evaluation during model selection.

\paragraph{Prompting setup.} In prompt-based LLM experiments, we evaluate zero-shot, 1-shot, and 3-shot prompting using the prompt templates in Appendix~\ref{sec:appendix Prompt templates}. For few-shot prompting, retrieved examples are selected from the training split only.

Vietnamese-input/Vietnamese-output NER uses Vietnamese demonstrations retrieved with \texttt{Vietnamese\_Embedding} \citep{Vietnamese_Embedding}.

English-input/Vietnamese-output NER uses parallel English-Vietnamese demonstrations retrieved by English-side dense similarity with \texttt{bge-large-en-v1.5} \citep{bge_embedding}.

The same number of demonstrations is used for all models under the same setting, and retrieved examples are inserted before the test input using the same JSON output format required from the model.

\paragraph{Open-source LLMs.} For the zero-shot and few-shot experiments, we use \texttt{Qwen2.5-3B-Instruct}, \texttt{Qwen2.5-7B-Instruct}, and \texttt{Qwen-SEA-LION-v4-8B-VL}. We use temperature \texttt{0.1}, maximum output length \texttt{2048}.

\paragraph{Closed-source LLMs.} For \texttt{gpt-5.4} and \texttt{gemini-3.1-flash-lite-preview}, we use temperature \texttt{1.0}, top-p \texttt{0.95}, and a maximum output length of \texttt{2,048} tokens. We set the reasoning/thinking level to \texttt{none} for \texttt{gpt-5.4} and \texttt{minimal} for \texttt{gemini-3.1-flash-lite-preview}. For both models, we allow up to two retries with a 3-second interval for transient API failures.

\paragraph{Hardware.} All fine-tuning experiments were run on two NVIDIA L40 48GB GPUs. 

\section{Other results}
\label{sec:appendix Other results}
This section reports additional results that support the main findings in the paper.

\begin{table}[!t]
\centering
\small
\setlength{\tabcolsep}{5pt}
\renewcommand{\arraystretch}{1.05}
\resizebox{0.8\columnwidth}{!}{%
\begin{tabular}{lrrrr}
\hline
\textbf{Tag} & \textbf{train} & \textbf{dev} & \textbf{test} & \textbf{Full Corpus} \\
\hline
T005 & 702 & 223 & 172 & 1,097 \\
T007 & 1,127 & 473 & 449 & 2,049 \\
T017 & 12,588 & 4,107 & 3,772 & 20,467 \\
T022 & 300 & 105 & 90 & 495 \\
T031 & 772 & 269 & 212 & 1,253 \\
T033 & 9,840 & 3,151 & 3,204 & 16,195 \\
T037 & 1,058 & 435 & 357 & 1,850 \\
T038 & 25,046 & 8,210 & 8,099 & 41,355 \\
T058 & 14,808 & 4,680 & 4,778 & 24,266 \\
T062 & 5,449 & 1,862 & 1,845 & 9,156 \\
T074 & 1,169 & 493 & 355 & 2,017 \\
T082 & 7,515 & 2,559 & 2,402 & 12,476 \\
T091 & 521 & 198 & 196 & 915 \\
T092 & 1,307 & 453 & 382 & 2,142 \\
T097 & 1,051 & 365 & 360 & 1,776 \\
T098 & 3,577 & 1,303 & 1,261 & 6,141 \\
T103 & 22,436 & 7,499 & 7,400 & 37,335 \\
T168 & 785 & 245 & 322 & 1,352 \\
T170 & 5,999 & 1,948 & 2,359 & 10,306 \\
T201 & 1,046 & 404 & 323 & 1,773 \\
T204 & 4,921 & 1,862 & 1,750 & 8,533 \\
\hline
\textbf{Total} & \textbf{122,017} & \textbf{40,844} & \textbf{40,088} & \textbf{202,949} \\
\hline
\end{tabular}%
}
\caption{Distribution of the 21 entity tags in the En-ViMedNER corpus across the train, development, test splits, and the full corpus.}
\label{tab:tag_distribution_envimedner}
\end{table}
\paragraph{Label distribution and frequency.} Table \ref{tab:tag_distribution_envimedner} shows that the corpus is not balanced across labels. Some tags, such as T038, T103, T058, and T017, appear much more often than others, while T022 and T091 are relatively rare. This imbalance reflects the natural distribution of biomedical concepts in MedMentions, making the task more realistic. However, it also means that models may learn frequent entity types more easily than rare ones.

\paragraph{Frequency does not fully explain performance.} To examine this effect more closely, Table \ref{tab:freq_semantic_types} compares the F1 scores of the five most and five least frequent semantic types. Label frequency does not fully determine F1: rare labels such as T031 (61.19) and T005 (59.07) exceed the macro-F1 of 49.01, while frequent T033 performs worst among the ten reported labels. This suggests that frequency helps learning, but semantic overlap and label confusability also substantially affect per-class performance. 

\begin{table}[t!]
\centering
\small
\begin{tabular}{llrr}
\toprule
\textbf{Frequency group} & \textbf{Label} & \textbf{Mentions} & \textbf{F1} \\
\midrule
\textbf{5 most frequent} \\
& T038 & 7,907 & 56.18 \\
& T103 & 7,247 & 68.30 \\
& T058 & 4,669 & 45.79 \\
& T017 & 3,676 & 51.95 \\
& T033 & 3,108 & 31.84 \\
\midrule
\textbf{5 least frequent} \\
& T168 & 317 & 41.46 \\
& T031 & 208 & 61.19 \\
& T091 & 191 & 36.68 \\
& T005 & 171 & 59.07 \\
& T022 & 87 & 37.21 \\
\bottomrule
\end{tabular}
\caption{F1 scores for the five most and least frequent semantic types.}
\label{tab:freq_semantic_types}
\end{table}

\begin{table*}[!t]
\centering
\small
\resizebox{\textwidth}{!}{%
\begin{tabular}{lcccccccc}
\hline
\textbf{Model} & \textbf{Params} & \textbf{Pretrain Data} & \textbf{Tokenization} & \textbf{Max Len} & \textbf{Precision} & \textbf{Recall} & \textbf{Micro-F1} & \textbf{Macro-F1}\\
\hline
\textbf{Vietnamese-supervised setting}\\
PhoBERT-base    & 135M & 20GB  & Word    & 256 & 52.65 & 49.98 & 51.28 & 47.22 \\
PhoBERT-base-v2  & 135M & 140GB & Word    & 256 & 52.60 & 50.42 & 51.49 & 46.48 \\
PhoBERT-large    & 370M & 20GB  & Word    & 256 & 52.19 & 50.84 & 51.50 & 48.89 \\
XLM-R-base      & 270M & 2.5TB & Syllable & 256 & 50.68 & 48.92 & 49.78 & 46.27 \\
XLM-R-large     & 550M & 2.5TB & Syllable & 256 & 52.66 & 50.29 & 51.45 & 48.80 \\
ViHealthBERT-word & 135M & 20GB+ & Word & 256 & 52.72 & 50.51 & 51.59 & 47.70\\
ViHealthBERT-syllable & 135M & 20GB+ & Syllable & 256 & 50.42 & 48.47 & 49.42 & 46.71\\
ViPubMedDeBERTa-xsmall & 22M & 158GB & Word & 256 & 52.35 & 52.74 & 52.54 & 48.67\\
ViPubMedDeBERTa-base & 86M & 158GB & Word & 256 & 52.86 & 52.54 & \textbf{52.70} & \textbf{49.01} \\

\hline
\textbf{English-supervised setting}\\
XLM-R-base & 270M & 2.5TB & Syllable & 256  & 27.80 & 35.48  & 31.12 & 27.10\\
XLM-R-large & 550M & 2.5TB & Syllable & 256  & 33.54 & 40.13  & 36.54 & 34.21\\
\hline
\textbf{Open-source LLMs}\\
Qwen2.5-3B-Instruct-0shot  & 3B & 18T & BPE & 2048 & 3.44 & 0.85 & 1.37  & 0.95\\
Qwen2.5-3B-Instruct-1shot  & 3B & 18T & BPE & 2048 & 8.87 & 2.48 & 3.87 & 3.09 \\
Qwen2.5-3B-Instruct-3shot  & 3B & 18T & BPE & 2048 & 11.04 & 2.66 & 4.29 & 3.29 \\
Qwen2.5-7B-Instruct-0shot  & 7B & 18T & BPE & 2048 & 2.66 & 0.33 & 0.59 & 0.67\\
Qwen2.5-7B-Instruct-1shot  & 7B & 18T & BPE & 2048 & 10.26 & 3.04 & 4.69 & 3.66 \\
Qwen2.5-7B-Instruct-3shot  & 7B & 18T & BPE & 2048 & 13.30 & 3.16 & 5.10 & 4.22\\
Qwen-SEA-LION-v4-8B-VL-0shot & 8B & 9M SFT pairs & BPE & 2048 & 10.57 & 1.57 & 2.73 & 2.46\\
Qwen-SEA-LION-v4-8B-VL-1shot & 8B & 9M SFT pairs & BPE & 2048 & 14.88 & 6.13 & 8.68 & 7.17 \\
Qwen-SEA-LION-v4-8B-VL-3shot & 8B & 9M SFT pairs & BPE & 2048 & 19.84 & 9.50 & 12.85 & 10.08 \\
\hline
\end{tabular}%
}
\caption{Detailed Vietnamese biomedical NER results on the full test set with Precision, Recall, Micro-F1, and Macro-F1. Encoder-based models are evaluated under Vietnamese-supervised and English-supervised settings, while open-source LLM prompting results are additionally reported on the full test set. Closed-source LLMs are omitted from this table because they are evaluated only on the mini-test set (due to API cost). \textit{Note:} For Qwen2.5 models, ``Pretrain Data'' denotes the reported 18T-token pretraining corpus of the Qwen2.5 series. For Qwen-SEA-LION-v4-8B-VL, ``Pretrain Data'' denotes the approximately 9M supervised fine-tuning instruction-text pairs used for SEA-LION regional adaptation.}
\label{tab:details}
\end{table*}

\paragraph{Model size is not the main factor.} Table~\ref{tab:details} provides detailed Vietnamese biomedical NER results on the full Vietnamese test set. The results show that model size alone does not determine performance. For example, \texttt{ViPubMedDeBERTa-base} achieves the best Micro-F1 despite being substantially smaller than \texttt{XLM-R-large}, while \texttt{ViPubMedDeBERTa-xsmall} remains competitive with much larger general-purpose encoders. This suggests that task- and domain-specific pretraining can be more important than model size for Vietnamese biomedical NER.

\section{Handling conflicts for BIO Tagging dataset} 

To convert the original entity annotations into token-level BIO labels, we first normalize span conflicts so that each token receives exactly one label. For nested entities, we keep only the outermost span, since it provides the most complete mention boundary for sequence labeling. 

For non-nested overlapping entities with different semantic labels, we do not automatically resolve the conflict because choosing one label would introduce additional ambiguity into the gold annotations. Instead, we log the corresponding file path and exclude the sentence from the current experimental setting. In total, this filtering removes 128 cases across the train, development, and test splits, accounting for less than 0.3\% of the full dataset.

\section{Postprocessing for LLMs experiments}

For LLM-based experiments, we apply a lightweight postprocessing step before evaluation to improve the validity and consistency of predicted BIO sequences. This step is used only for LLM prompting outputs, since LLMs generate free-form responses and may produce malformed JSON, missing fields, or invalid BIO transitions. We do not apply the same postprocessing to fine-tuned encoder models, which are trained and decoded directly as token-level sequence taggers over a fixed BIO label space. Applying extra rule-based or LLM-based corrections to these models would introduce additional inference-time supervision, making the comparison less direct.

We keep "các" and "những" inside an entity span when adjacent words are labeled as part of the entity, but remove "khác" and "này" because they usually act as modifiers rather than core entity tokens. We also remove "mỗi" unless the corresponding English entity begins with \texttt{per}, to avoid incorrectly including mistranslated rate or frequency expressions.

Finally, we repair invalid BIO transitions such as \texttt{O-I} and \texttt{B-X I-Y}. For simple \texttt{O-I} cases, the invalid \texttt{I} tag is converted into a valid beginning tag when appropriate. For inconsistent adjacent entity types, we use the prompt template in Figure~\ref{fig:bio_conflict_resolution_prompt} to select the most plausible entity type from the sentence context and tag definitions.

\section{Prompt templates}
\label{sec:appendix Prompt templates}

This appendix presents the prompt templates used in this study.

\begin{figure*}[t]
\centering
\begin{tcolorbox}[
   width=1\textwidth,
   colback=white,
   colframe=black,
   boxrule=0.8pt,
   arc=2pt,
   left=3pt,
   right=3pt,
   top=3pt,
   bottom=3pt,
   title=\textbf{Prompt template for BIO conflict resolution},
   coltitle=white,
   colbacktitle=black,
   fonttitle=\bfseries
]
\begin{lstlisting}[style=promptstyle]
You are a biomedical named entity expert.

# TASK
You are given a biomedical BIO tagging conflict. Resolve the conflict.

# INPUT

Sentence:
{sentence}

Tokens:
{json.dumps(tokens, ensure_ascii=False)}

Current labels:
{json.dumps(labels, ensure_ascii=False)}

Conflict:
- token[{i}] = "{tokens[i]}" has label "{labels[i]}"
- token[{j}] = "{tokens[j]}" has label "{labels[j]}"
This is invalid because B-X cannot be followed immediately by I-Y when X != Y.

Choose the single best entity type for this local span based on the sentence context.

# TAG DEFINITIONS
{tag_legend}

Return JSON only:
{{
  "resolved_type": "<one tag type such as T033>",
  "tokens": ["<token1>", "<token2>", "<token3>"],
  "labels": ["<label1>", "<label2>", "<label3>"]
}}
\end{lstlisting}
\end{tcolorbox}
\caption{Prompt template used for resolving invalid BIO transitions in biomedical named entity recognition outputs.}
\label{fig:bio_conflict_resolution_prompt}
\end{figure*}

\begin{figure*}[!t]
\centering
\begin{tcolorbox}[
   width=1\textwidth,
   colback=white,
   colframe=black,
   boxrule=0.8pt,
   arc=2pt,
   left=3pt,
   right=3pt,
   top=3pt,
   bottom=3pt,
   title=\textbf{Prompt template for few-shot Vietnamese NER experiments},
   coltitle=white,
   colbacktitle=black,
   fonttitle=\bfseries
]
\begin{lstlisting}[style=promptstyle]
You are a biomedical named entity recognition expert.

# TASK
You are given one Vietnamese biomedical sentence. Perform token-level biomedical named entity recognition using the BIO tagging scheme.

Your job is to:
1. split the sentence into tokens,
2. assign exactly one label to each token,
3. use BIO tagging:
- B-<TAG> for the beginning of an entity,
- I-<TAG> for the continuation of the same entity,
- O for a token outside any entity.

# IMPORTANT RULES
- Every token must receive exactly one label.
- Preserve the original sentence content.
- Do not add, remove, paraphrase, or translate words.
- Use only the semantic tags defined below.
- If a token is not part of a biomedical named entity, label it O.
- If an entity has multiple tokens, the first token must be B-<TAG> and the remaining tokens must be
I-<TAG>.
- Do not create overlapping entities.
- Learn the JSON output style from the examples below.

# TAG DEFINITIONS
{tag_legend}

# ANSWER FORMAT
Return JSON only:
{{
  "vi_text": "<original Vietnamese sentence>",
  "tokens": ["<token1>", "<token2>", "<token3>"],
  "labels": ["<label1>", "<label2>", "<label3>"]
}}

Constraints for output JSON:
- "tokens" and "labels" must have the same number of items.
- Each label in "labels" must be exactly one of: O, B-<TAG>, or I-<TAG> where <TAG> is from 
TAG DEFINITIONS.
- Keep "vi_text" unchanged from the input sentence.

# FEW-SHOT EXAMPLES
{example_text}

# INPUT SENTENCE
{input_text}
\end{lstlisting}
\end{tcolorbox}
\caption{Prompt template used for few-shot Vietnamese NER experiments. For zero-shot experiments, the section \textbf{\# FEW-SHOT EXAMPLES} is excluded.}
\label{fig:fewshot_vi_vi_prompt}
\end{figure*}

\begin{figure*}[t]
    \centering
    \includegraphics[width=0.9\textwidth]{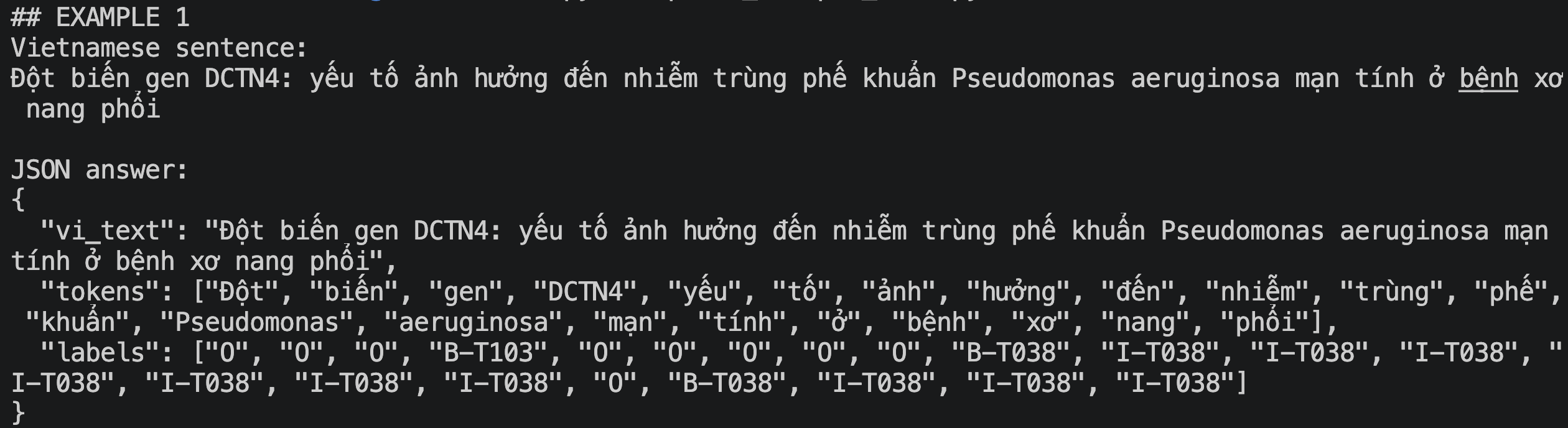}
    \caption{Example of \textbf{example\_text} in one-shot Vietnamese NER prompt.}
    \label{fig:prompt-template-vivi}
\end{figure*}

\begin{figure*}[!t]
\centering
\begin{tcolorbox}[
   width=1\textwidth,
   colback=white,
   colframe=black,
   boxrule=0.8pt,
   arc=2pt,
   left=3pt,
   right=3pt,
   top=3pt,
   bottom=3pt,
   title=\textbf{Prompt template for few-shot cross-lingual NER experiments},
   coltitle=white,
   colbacktitle=black,
   fonttitle=\bfseries
]
\begin{lstlisting}[style=promptstyle]
You are a biomedical named entity recognition expert for English-to-Vietnamese transfer.

# TASK
Your job is to:
1. translate the input sentence into Vietnamese in medical domain with medical terms,
2. split the Vietnamese sentence into tokens,
3. assign exactly one BIO label to each Vietnamese token.
4. Use BIO tagging:
- B-<TAG> for the beginning of an entity,
- I-<TAG> for the continuation of the same entity,
- O for a token outside any entity.

# IMPORTANT RULES
- The output sentence must be in Vietnamese.
- The meaning must stay faithful to the English input.
- The tokens and labels must correspond to the Vietnamese sentence only.
- Every token must receive exactly one label.
- Use only the semantic tags defined below.
- If a token is not part of a biomedical named entity, label it O.
- If an entity has multiple tokens, the first token must be B-<TAG> and the remaining tokens must be 
I-<TAG>.
- Do not create overlapping entities.

# TAG DEFINITIONS
{tag_legend}

# ANSWER FORMAT
Return JSON only:
{{
  "vi_text": "<Vietnamese translation>",
  "tokens": ["<token1>", "<token2>", "<token3>"],
  "labels": ["<label1>", "<label2>", "<label3>"]
}}

# FEW-SHOT EXAMPLES
{example_text}

# INPUT ENGLISH SENTENCE
{input_text}
\end{lstlisting}
\end{tcolorbox}
\caption{Prompt template used for few-shot cross-lingual NER experiments. For zero-shot experiments, the section \textbf{\# FEW-SHOT EXAMPLES} is excluded.}
\label{fig:fewshot_en_vi_prompt}
\end{figure*}

\begin{figure*}[!t]
    \centering
    \includegraphics[width=0.9\textwidth]{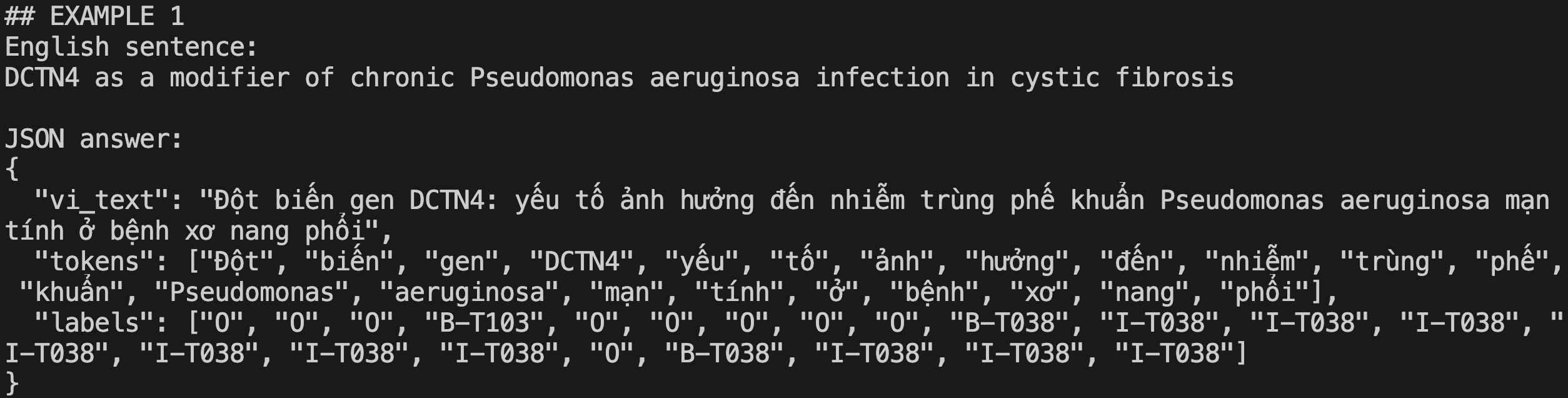}
    \caption{Example of \textbf{example\_text} in one-shot cross-lingual NER prompt.}
    \label{fig:prompt-template-envi}
\end{figure*}

\begin{figure*}[!t]
\centering
\begin{tcolorbox}[
   width=1\textwidth,
   colback=white,
   colframe=black,
   boxrule=0.8pt,
   arc=2pt,
   left=3pt,
   right=3pt,
   top=3pt,
   bottom=3pt,
   title=\textbf{Prompt template for LLM-as-a-judge evaluation of cross-lingual NER outputs},
   coltitle=white,
   colbacktitle=black,
   fonttitle=\bfseries
]
\begin{lstlisting}[style=promptstyle]
Given the prediction tagged:
"{pred_sent}"

Prediction tagged entities list:
"{pred_entities_list}"

Given ground truth:
"{gt_sent}"

Ground truth entities list:
"{gt_entities_list}"

For each ground truth entity, find the best matching predicted entity, allowing paraphrases. Each predicted entity can be used at most once.

Rules:
- "pred text label" = "correct" if pred text matches gt meaning; otherwise "incorrect"; use "miss" if no match.
- "pred tag label" = "correct" if pred tag = gt tag; otherwise "incorrect"; use "miss" if no match.
- If no match, set "pred text" = "" and "pred tag" = "".
- Return unmatched predicted entities separately.

Return ONLY valid JSON parseable by json.loads() as:
{
  "matches": [
   {
    "gt text": "...",
    "gt tag": "...",
    "pred text": "...",
    "pred text label": "correct|incorrect|miss",
    "pred tag": "...",
    "pred tag label": "correct|incorrect|miss"
   }
  ],
  "unmatched": [
   {
    "pred text": "...",
    "pred tag": "..."
   }
  ]
}

Do not output markdown, code fences, explanations, or any text outside the JSON object.
\end{lstlisting}
\end{tcolorbox}
\caption{Prompt template used for LLM-as-a-judge evaluation of cross-lingual NER outputs.}
\label{fig:llm_judge_prompt}
\end{figure*}

\end{document}